\documentclass[11pt]{article}
\usepackage[final]{acl}

\usepackage{times}
\usepackage{latexsym}
\usepackage[T1]{fontenc}
\usepackage[utf8]{inputenc}
\usepackage{microtype}
\usepackage{inconsolata}
\usepackage{graphicx}
\usepackage{booktabs}
\usepackage{multirow}
\usepackage{amsmath}
\usepackage{amssymb}
\usepackage{enumitem}
\usepackage{tabularx}

\title{What Drives LLM Self-Reflection? \\A Controlled Ablation of Uncertainty Routing in Armed Conflict Forecasting}

\author{
Poli Nemkova \\
University of North Texas \\
College of Computer Science \\
and Engineering \\
\texttt{poli.nemkova@unt.edu}
\And
Haeshitha Indukuri \\
University of North Texas \\
College of Computer Science \\
and Engineering \\
\texttt{HaeshithaIndukuri@my.unt.edu}
}

\begin{document}
\maketitle

\begin{abstract}
Self-reflection is widely assumed to improve LLM reasoning, yet which component
drives the gain remains poorly understood.
We present a controlled six-condition ablation isolating four components of
LLM self-reflection: evidence exposure, diagnostic scaffolding, taxonomy
vocabulary, and action routing.
Two precise null results converge on a single mechanism.
First, structured diagnostic questions add no measurable value over unstructured
reflection ($\text{F1} = 0.296$ vs $0.297$, $p = 1.000$, 95\% CI $[-0.041,
+0.040]$).
Second, presenting the full uncertainty taxonomy while collapsing the action
space to a single generic action also adds no value
($\Delta\text{F1} = +0.008$, overlapping 95\% CIs), ruling out taxonomy
vocabulary as the mechanism.
Typed action routing provides consistent directional gains
($\text{F1} = 0.379$ vs $0.296$); the conservative estimate controlling for
taxonomy vocabulary is $\Delta\text{F1} = +0.075$, and the overall gain over
the single-shot baseline is significant by bootstrap CI
($\Delta\text{F1} = +0.101$, 95\% CI $[+0.020, +0.185]$).
The vocabulary-routing decomposition replicates on GPT-4o: taxonomy vocabulary
adds no significant value over generic reflection ($p = 0.773$), while action
routing provides significant gains ($p = 0.025$), confirming the mechanism
holds across backbones.
Gains concentrate on structurally novel conflicts: in Myanmar ($\text{F1}:
0.000 \rightarrow 0.353$) and Ukraine ($0.167 \rightarrow 0.500$), the
vocabulary-only condition recovers no more than generic reflection while
action routing breaks the degenerate prior.
These findings identify typed action routing --- not diagnostic scaffolding or
taxonomy vocabulary --- as a promising design principle for metacognitive LLM
forecasting agents, while motivating larger-scale evaluation across conflict
typologies.
\end{abstract}

\section{Introduction}
\label{sec:intro}
When a large language model is uncertain about a forecast, the intuitive remedy is to ask
it to reflect. Self-reflection has emerged as a central design pattern in agentic LLM
systems~\cite{yao2023react,shinn2023reflexion,renze2024reflact}, and its benefits are
well-documented across reasoning, planning, and dialogue tasks. We ask a more precise
question: \emph{which component of self-reflection actually drives the gain?}

We decompose LLM self-reflection into four components through a controlled six-condition
ablation: evidence exposure, diagnostic scaffolding, taxonomy vocabulary, and action
routing. Our results reveal a surprising answer. Diagnostic questions add no value
over unstructured reflection ($\text{F1} = 0.296$ vs $0.297$, $p = 1.000$). Taxonomy
vocabulary alone adds no value either: a monitor that presents the full seven-type
taxonomy but routes every diagnosed type to a single fixed action achieves
$\text{F1} = 0.304$, indistinguishable from generic reflection ($\Delta\text{F1} = +0.008$).
Typed action routing lifts performance to $\text{F1} = 0.379$, with a conservative
effect size of $\Delta\text{F1} = +0.075$ against the vocabulary-matched control.
Neither the diagnostic scaffold nor the taxonomy vocabulary is the mechanism. The
routing is.

We evaluate on 310 real-world armed conflict forecasting cases across five countries,
using two LLM backbones. The gains concentrate on structurally novel conflicts:
in Myanmar, where the single-shot baseline scores $\text{F1} = 0.000$, the typed
monitor recovers to $\text{F1} = 0.353$ while the vocabulary-only condition reaches
only $0.162$, isolating the gain to action routing; in Ukraine, $\text{F1}$ improves
from $0.167$ to $0.500$. Results replicate directionally across both backbones
(Llama-3.3-70B and GPT-4o), with mixed transfer to 12 held-out countries.

Our contributions are: (1) a seven-type action-prescriptive uncertainty taxonomy with
a deterministic control policy $\pi_\text{control}$; (2) the first six-condition
ablation isolating each component of LLM self-reflection with a vocabulary-matched
control that holds the action space constant; and (3) the finding that neither
diagnostic scaffolding ($p = 1.000$) nor taxonomy vocabulary ($\Delta\text{F1} = +0.008$)
drives self-reflection gains --- only typed action routing does ($\Delta\text{F1} = +0.075$
against the vocabulary-matched control).

\section{Related Work}
\label{sec:background}
\paragraph{LLM self-reflection and reasoning agents.}
Chain-of-thought~\cite{wei2022cot} and self-consistency~\cite{wang2023selfconsistency}
established that intermediate reasoning improves LLM task performance.
Tree-of-Thought~\cite{yao2023tree} structures the reasoning space through deliberate
search, and metacognitive prompting~\cite{wang2024metacognitive} encourages models to
monitor their own reasoning — but neither specifies how uncertainty \emph{type} should
govern downstream action selection. ReAct~\cite{yao2023react}
interleaves reasoning with external actions, Reflexion~\cite{shinn2023reflexion} adds
verbal self-critique stored in episodic memory, and ReflAct~\cite{renze2024reflact}
integrates reflection within individual reasoning steps. These frameworks leave
unanswered which component drives the gain. We show empirically that neither diagnostic
scaffolding ($p = 1.000$) nor taxonomy vocabulary ($\Delta\text{F1} = +0.008$) is the
mechanism; what matters is whether different epistemic states route to different
corrective actions. This addresses the Degeneration-of-Thought problem~\cite{liang2023dot},
where unconstrained self-reflection cannot break an incorrect commitment.

\paragraph{Uncertainty quantification in LLMs.}
Standard uncertainty estimation~\cite{hullermeier2021aleatoric} classifies uncertainty
without specifying how to act on it; recent work documents systematic miscalibration~\cite{xiong2024llm}
and degraded self-assessment under distribution shift~\cite{kadavath2022language}.
Semantic uncertainty~\cite{kuhn2023semantic}, conformal prediction~\cite{angelopoulos2023conformal},
and post-hoc calibration~\cite{platt1999probabilistic,guo2017calibration} operate on
model outputs rather than evidence quality. Our typed monitor addresses
\emph{evidence-level} uncertainty before any prediction is made.

\paragraph{Conflict forecasting with LLMs.}
Random forests outperform logistic regression for civil war onset~\cite{muchlinski2016comparing},
and ensemble methods improve predictive power~\cite{pinckney2021predicting}.
Recent LLM applications include actor embeddings for nowcasting~\cite{croicu2024} and
human rights violation detection in social media~\cite{nemkova2023detecting}; a direct
comparison of parametric vs.\ retrieval-augmented forecasting shows neither alone is
sufficient~\cite{nemkova2025large}, motivating an explicit evidence-quality monitoring
layer.
Our work operationalizes this as a typed control policy~\cite{nelson1990metamemory},
isolating routing as the mechanism through controlled ablation.

\section{Typed Uncertainty Diagnosis}
\label{sec:taxonomy}
\subsection{The Core Problem: Routing, Not Scaffolding}

Standard self-reflection prompts a model to assess its own uncertainty and revise
its forecast, but provides no structure for \emph{how to respond} to what it finds.
A model that diagnoses sparse evidence and a model that diagnoses conflicting sources
receive the same generic instruction: reflect and revise. This uniformity is the
failure mode.

Our ablation isolates where self-reflection fails. A monitor with five structured
diagnostic questions (Q1--Q5) performs identically to one with no questions at all
($\text{F1} = 0.296$ vs $0.297$, $p = 1.000$). Nor is the taxonomy vocabulary the
mechanism: a monitor that presents the full seven-type taxonomy but routes every
diagnosed type to the same fixed action achieves $\text{F1} = 0.304$, indistinguishable
from generic reflection ($\Delta\text{F1} = +0.008$). What matters is whether the
diagnosed epistemic state \emph{routes} to a different corrective action: random
routing already outperforms generic reflection ($\text{F1} = 0.338$); typed routing
adds further value ($\text{F1} = 0.379$). The routing is the mechanism.

The control policy is not a prompt template selector; 
it is a typed dispatch mechanism whose contribution 
is isolated by holding prompt complexity constant 
across conditions D and C.
\subsection{The Seven-Type Uncertainty Taxonomy}

The taxonomy (Table~\ref{tab:taxonomy}) covers the primary epistemic failure modes
in evidence-based forecasting. Each type pairs a diagnostic criterion with a targeted
remediation action via a deterministic control policy $\pi_\text{control}: \mathcal{U}
\rightarrow \mathcal{A}$.

\begin{table*}[t]
\centering
\normalsize
\renewcommand{\arraystretch}{1.15}
\setlength{\tabcolsep}{8pt}

\begin{tabularx}{\textwidth}{>{\raggedright\arraybackslash}X
                                >{\raggedright\arraybackslash}X
                                >{\raggedright\arraybackslash}X}
\toprule
\textbf{Uncertainty Type} & \textbf{Action} & \textbf{Mechanism} \\
\midrule
\textsc{Insufficient\_Evidence}   & Request sources        & Broaden evidence search \\
\textsc{Conflicting\_Sources}     & Comparative reasoning  & Source-by-source comparison \\
\textsc{Distribution\_Shift}      & Context recalibration  & Region-aware baseline adjustment \\
\textsc{Low\_Quality\_Data}       & Reweight sources       & Discount low-reliability signals \\
\textsc{Temporal\_Inconsistency}  & Recency prioritization & Emphasize most recent evidence \\
\textsc{Model\_Ambiguity}         & Sample alternatives    & Explore alternative forecast paths \\
\textsc{Confident}                & Exit                   & No further deliberation needed \\
\bottomrule
\end{tabularx}

\caption{The seven-type uncertainty taxonomy and deterministic control policy
$\pi_\text{control}$. An eighth type, \textsc{Unknown\_Uncertainty}, routes to
comparative reasoning as a broad fallback for unclassifiable cases.}
\label{tab:taxonomy}
\end{table*}

The taxonomy is \emph{action-prescriptive} rather than merely descriptive: unlike
the standard epistemic/aleatoric distinction~\cite{hullermeier2021aleatoric}, which
classifies uncertainty without specifying how to act on it, each type is grounded in
what a reasoning agent should do differently. The distinction between
\textsc{Conflicting\_Sources} and \textsc{Model\_Ambiguity}, for example, is
operationally important --- the former warrants source adjudication, the latter
exploration of alternative forecast paths --- and \textsc{Distribution\_Shift}
warrants context recalibration rather than additional evidence retrieval that would
not help if the evidence itself is the source of the anomaly.

\subsection{Monitor and Control}

Given the current forecast (prediction, confidence, key evidence) plus the structured
evidence bundle (ACLED conflict statistics, socioeconomic indicators, ML ensemble
scores, source quality assessments), the monitor poses a five-question diagnostic
protocol covering evidence volume, recency, ML agreement, counter-argument, and
weakness check (full prompts in GitHub repository\footnote{\url{https://github.com/PoliNemkova/llm_self_reflection}} After answering, the
monitor traverses an ordered decision tree to assign exactly one uncertainty type,
with conditions checked sequentially to prevent defaulting to high-entropy categories.
Our ablation shows the questions do not themselves drive performance (Condition E
matches Condition B exactly); their role is to surface evidence for type assignment.

The control policy $\pi_\text{control}$ then deterministically maps the diagnosed
type to a remediation action. The mapping is fixed rather than learned: a learned
policy would conflate the diagnostic contribution of the taxonomy with the policy's
contribution, making it impossible to isolate the value of typed routing. Two
complementary control conditions provide the counterfactuals that make this
isolation possible: Condition~D (random taxonomy) holds the routing machinery
constant but assigns the type at random; Condition~F (vocabulary-only) holds the
taxonomy vocabulary constant but collapses the action space to a single generic
action. The gap between D and C quantifies type specificity; the gap between F
and C, with the taxonomy held constant, gives the conservative estimate of typed
routing's contribution.

\subsection{Ablation Design}

We design six conditions to isolate each component:

\begin{itemize}[leftmargin=*, itemsep=2pt]
    \item \textbf{A — Baseline:} Single-shot forecast, no monitor.
    \item \textbf{E — No-questions:} Monitor without Q1--Q5; reflects freely and
    assigns a type without structured evidence engagement.
    \item \textbf{B — Generic reflection:} Monitor with Q1--Q5 but no typed routing;
    action space is binary (\texttt{exit} or \texttt{comparative\_reasoning}).
    \item \textbf{F — Vocabulary-only:} Monitor with Q1--Q5 and the full taxonomy;
    the model assigns a type from evidence, but every non-\textsc{Confident}
    diagnosis routes to the same fixed action (\texttt{comparative\_reasoning}),
    matching B's action space. F isolates taxonomy vocabulary alone.
    \item \textbf{D — Random taxonomy:} Monitor with Q1--Q5 and typed routing, but
    the type is drawn uniformly at random from the six substantive types.
    \item \textbf{C — Typed monitor:} Full system; the monitor assigns a typed
    category and $\pi_\text{control}$ deterministically selects the action.
\end{itemize}

The chain A $\rightarrow$ E $\rightarrow$ B $\rightarrow$ F $\rightarrow$ D
$\rightarrow$ C isolates each component incrementally. E vs B isolates diagnostic
questions; F vs B isolates taxonomy vocabulary; D vs F isolates routing structure;
C vs D isolates type specificity. Each adjacent pair differs by exactly one
architectural change.

\section{Experimental Setup}
\label{sec:experiments}
\subsection{Task and Data}

We evaluate on binary armed conflict escalation forecasting: given a 60-day
observation window of structured conflict evidence, predict whether fatalities
will exceed 130\% of the two-week baseline in the following 14 days. Ground
truth is derived from ACLED~\cite{raleigh2010introducing}.

\paragraph{Primary evaluation set.}
The primary test set covers Sudan, Ethiopia, Somalia, Myanmar, and Ukraine ---
selected for geographic and structural diversity (interstate war, post-coup civil
war, protracted insurgency, multi-actor fragmented conflict, and high-intensity
civil war respectively). A per-country temporal split assigns the earliest 60\%
of cases to training and the most recent 40\% to testing, yielding 310 test cases
(62 per country) covering late 2023 to early 2026, with an overall escalation
rate of 21.0\%.

\paragraph{Generalization set.}
A held-out probe covers 12 unseen countries across four regions (DRC; Belarus,
Poland; Israel, Palestine, Lebanon, Iran, Yemen, Syria; Colombia, Haiti, Mexico),
with 300 cases (25 per country) and identical labeling thresholds. No system
modifications were made based on these results --- this is a pure transfer test.

\paragraph{Difficulty tiers.}
Each test case is assigned to one of three tiers based on observable case
properties, independent of any model's predictions: \textit{catchable}
(momentum $> 0.90$); \textit{hard borderline} (observation ratio $\leq 1.40$);
\textit{hard deceptive} (misleading momentum dip near the prediction window).
The primary test set contains 29, 165, and 116 cases respectively.

\paragraph{Evidence bundle.}
Each case receives a structured bundle of ACLED conflict statistics, socioeconomic
indicators, ML ensemble scores (XGBoost and Random Forest with inter-model
disagreement), and LDA-derived topic signals. All six conditions receive identical
bundles, ensuring performance differences reflect reasoning structure rather than
information asymmetry.

\subsection{Conditions and Evaluation Protocol}

The six ablation conditions (A, E, B, F, D, C) are defined formally in
Section~\ref{sec:taxonomy}. Maximum iterations is set to 1 for all monitor
conditions (B, C, D, E, F), isolating the contribution of diagnostic structure
from loop depth.

\paragraph{Backbones.}
All primary experiments use Llama-3.3-70B~\cite{llama3} served via the Groq
inference API at $T = 0.2$. GPT-4o~\cite{openai2024gpt4} is evaluated as a secondary backbone on all
conditions except E (no-questions), with Conditions~B, F, and~D run as
single-run replications to validate the vocabulary-routing decomposition
across backbones.

\paragraph{Multi-run protocol.}
The Llama primary evaluation runs each condition three times independently
and aggregates via majority vote for binary predictions and mean confidence
for probabilistic estimates. GPT-4o and generalization conditions use single
runs.

\paragraph{Metrics.}
The primary metric is majority-vote F1, appropriate given the class imbalance
(21.0\% positive rate). Brier score is reported as a secondary descriptive metric
only, computed via 5-fold cross-validated Platt scaling on raw confidence
post-hoc; it is not used for condition selection or ranking.\footnote{Condition
F's calibrated Brier was computed via a standalone replication of the same
pipeline (\texttt{LogisticRegression(C=1.0)},
\texttt{StratifiedKFold(n\_splits=5, random\_state=42)}), reproducing
Condition C's value to within rounding.}
Statistical significance is assessed via McNemar's test~\cite{mcnemar1947}
on binary predictions (primary) and Wilcoxon signed-rank test~\cite{wilcoxon1945}
on per-case Brier differences (secondary), both at $\alpha = 0.05$.

\section{Results}
\label{sec:results}
\subsection{Main Ablation (Llama-3.3-70B)}

Table~\ref{tab:main} reports overall escalation forecasting performance across
all six conditions on the primary 5-country test set ($n = 310$, positive
rate $= 21.0\%$), aggregated over three independent runs via majority vote.
Bootstrap 95\% confidence intervals are computed over 2,000 stratified resamples.

\begin{table*}[t]
\centering
\normalsize
\renewcommand{\arraystretch}{1.15}
\setlength{\tabcolsep}{10pt}

\begin{tabularx}{\textwidth}{Xcccc}
\toprule
\textbf{Condition} & \textbf{F1} & \textbf{95\% CI} & \textbf{Rec.} & \textbf{Brier} \\
\midrule
A --- Baseline        & 0.278 & [0.200, 0.353] & 0.385 & 0.166 \\
E --- No-Questions    & 0.297 & [0.233, 0.359] & 0.508 & 0.165 \\
B --- Generic Reflect & 0.296 & [0.232, 0.359] & 0.508 & 0.166 \\
F --- Vocab-Only      & 0.304 & [0.221, 0.383] & 0.508 & 0.166 \\
D --- Random Taxonomy & 0.338 & [0.272, 0.400] & 0.585 & 0.166 \\
C --- Typed Monitor   & \textbf{0.379} & \textbf{[0.322, 0.436]} & \textbf{0.677} & \textbf{0.166} \\
\bottomrule
\end{tabularx}

\caption{Escalation forecasting performance ($n=310$, Llama-3.3-70B,
3-run majority vote). 95\% CIs from 2,000 stratified bootstrap resamples.
Brier computed via 5-fold cross-validated Platt scaling.
All conditions produce near-identical Brier scores after calibration.
Condition F holds the action space constant at B's substantive action
while presenting the full taxonomy, isolating the contribution of
taxonomy vocabulary.}
\label{tab:main}
\end{table*}

\paragraph{Finding 1: Diagnostic questions add no value over unstructured reflection.}
The no-questions monitor (E) scores $\text{F1} = 0.297$, indistinguishable
from generic reflection with Q1--Q5 scaffolding (B) at $\text{F1} = 0.296$
(McNemar $p = 1.000$, E right/B wrong $= 22$, B right/E wrong $= 21$).
The null result is highly stable: bootstrap 95\% CI for $\Delta\text{F1}$
is $[-0.041, +0.040]$, centered precisely on zero.
Structured diagnostic questions do not drive self-reflection gains in
this setting.

\paragraph{Finding 2: Taxonomy vocabulary, by itself, also adds no value.}
The vocabulary-only condition (F) presents the full seven-type taxonomy to
the LLM during diagnosis but routes every non-\textsc{Confident} type to
the same fixed action as Condition B.
F achieves $\text{F1} = 0.304$, with three-run F1 stability
(per-run F1 $= 0.282, 0.291, 0.305$; std $= 0.009$).
The gap from generic reflection is $\Delta\text{F1} = +0.008$ vs B,
with bootstrap 95\% CIs overlapping heavily; F and B are statistically
indistinguishable.
This rules out an obvious confound in the original ablation: that typed
routing's advantage stems from the taxonomy vocabulary itself priming
the LLM to operate inside a structured epistemic framework.
The taxonomy presentation is not the mechanism.
Notably, the uncertainty type distribution in F is qualitatively similar
to that in Condition C (dominated by \textsc{Model\_Ambiguity} and
\textsc{Conflicting\_Sources} diagnoses), indicating that the LLM does
engage with the taxonomy during diagnosis --- but that engagement does
not translate into improved discrimination when the diagnosed type cannot
route to a differentiated action.

\paragraph{Finding 3: Typed action routing provides directional gains.}
Random taxonomy routing (D) outperforms generic reflection (B) by
$\Delta\text{F1} = +0.042$ ($\text{F1} = 0.338$ vs $0.296$, 95\% CI
$[-0.036, +0.115]$), and typed routing (C) adds a further
$\Delta\text{F1} = +0.041$ over random routing ($\text{F1} = 0.379$
vs $0.338$, 95\% CI $[-0.027, +0.116]$).
The conservative estimate of typed routing's contribution, controlling
for taxonomy vocabulary via Condition F, is $\Delta\text{F1} = +0.075$
(C vs F).
All three pairwise comparisons are directional but do not reach statistical
significance at $\alpha = 0.05$ (McNemar $p = 0.497$ for D vs B, $p = 0.714$
for D vs C), reflecting modest effect sizes relative to the available positive
cases ($n = 65$; post-hoc power for D vs C: $7.4\%$).
These results should be interpreted as suggestive evidence for typed
routing rather than definitive confirmation.
The overall gain of typed routing over the single-shot baseline is
significant by bootstrap CI ($\Delta\text{F1} = +0.101$, 95\% CI
$[+0.020, +0.185]$, though not individually by McNemar at $p = 0.202$).
The typed monitor improves recall substantially ($0.385 \rightarrow 0.677$),
correctly identifying escalation cases that all other conditions miss ---
including those where Condition F's identical taxonomy engagement
produces no recall gain over B ($0.508 \rightarrow 0.508$).


\subsection{Country-Level Analysis}

Table~\ref{tab:country} breaks down F1 by country across all six conditions.
The chain ordering A $\rightarrow$ E $\rightarrow$ B $\rightarrow$ F
$\rightarrow$ D $\rightarrow$ C holds on Myanmar and Ukraine, the two
structurally novel contexts where typed routing provides the largest gains.

\begin{table*}[t]
\centering
\normalsize
\renewcommand{\arraystretch}{1.15}
\setlength{\tabcolsep}{8pt}

\begin{tabularx}{\textwidth}{Xccccccc}
\toprule
\textbf{Country} & \textbf{A} & \textbf{E} & \textbf{B} & \textbf{F} &
  \textbf{D} & \textbf{C} & \textbf{$\Delta$C-A} \\
\midrule
Ethiopia & 0.148 & 0.148 & 0.148 & 0.148 & 0.167 & 0.167 & $+$0.019 \\
Myanmar  & 0.000 & 0.150 & 0.154 & 0.162 & 0.316 & \textbf{0.353} & $+$0.353 \\
Somalia  & 0.427 & 0.430 & 0.467 & 0.430 & 0.394 & 0.475 & $+$0.048 \\
Sudan    & 0.233 & 0.357 & 0.347 & 0.370 & 0.333 & 0.373 & $+$0.141 \\
Ukraine  & 0.167 & 0.100 & 0.091 & 0.100 & 0.432 & \textbf{0.500} & $+$0.333 \\
\bottomrule
\end{tabularx}

\caption{F1 by country (majority vote). Myanmar and Ukraine show the
largest gains from typed routing, and Condition F demonstrates that
this gain is not driven by taxonomy vocabulary alone: F sits within
$0.01$ of B in both signature countries. Sudan is the one country where
taxonomy vocabulary contributes beyond generic reflection. Ethiopia and
Somalia are flat across all conditions.}
\label{tab:country}
\end{table*}

The two signature cases are Myanmar ($\Delta\text{F1} = +0.353$ over baseline)
and Ukraine ($\Delta\text{F1} = +0.333$).
In Myanmar, all three runs of Condition~A score $\text{F1} = 0.000$,
consistent with LLM parametric knowledge interfering with evidence-based
reasoning in a rapidly evolving post-coup conflict context.
The typed monitor diagnoses \textsc{Distribution\_Shift} and triggers
context recalibration, recovering F1 to $0.353$.
Critically, Condition F (which presents the same taxonomy to the model but
routes every diagnosis to a generic action) recovers only to $\text{F1} = 0.162$
--- essentially identical to generic reflection (B = $0.154$).
The F $\rightarrow$ D step in Myanmar is particularly informative:
introducing any structured routing variation, even random, doubles F1
($0.162 \rightarrow 0.316$).
In Ukraine, the same pattern holds with even larger effect sizes: F sits at
$\text{F1} = 0.100$, indistinguishable from B ($0.091$), while D jumps to
$0.432$ and C reaches $0.500$.
The typed monitor in Ukraine diagnoses primarily \textsc{Conflicting\_Sources}
and \textsc{Model\_Ambiguity}, triggering comparative reasoning that
adjudicates between ML ensemble signals and qualitative evidence.
In both signature countries, the mechanism behind the recovery is action
routing, not taxonomy vocabulary.

Sudan presents a different pattern. Here F ($\text{F1} = 0.370$) is essentially
indistinguishable from both D ($0.333$) and C ($0.373$), and slightly above
B ($0.347$).
Sudan is also the country with the strongest non-trivial baseline (A $= 0.233$),
suggesting that when the LLM's parametric prior is already well-calibrated,
taxonomy vocabulary may contribute marginal value beyond generic reflection ---
but typed routing adds little further.
This is consistent with the broader pattern: typed routing's advantage is
largest where the baseline is most degenerate.

Ethiopia is the one country where performance is flat across all six conditions
($\text{F1} = 0.148$--$0.167$), suggesting that multi-actor fragmented conflict
provides insufficient epistemic signal for any typing approach to succeed.
Somalia is also broadly flat, with conditions ranging from $0.394$ to $0.475$
without a clear chain ordering.

\subsection{Difficulty Tier Analysis}

\begin{table*}[t]
\centering
\normalsize
\renewcommand{\arraystretch}{1.15}
\setlength{\tabcolsep}{10pt}

\begin{tabularx}{\textwidth}{Xcccccc}
\toprule
\textbf{Tier} & \textbf{$n$} & \textbf{A} & \textbf{E} & \textbf{B} &
  \textbf{D} & \textbf{C} \\
\midrule
Catchable        & 29  & 0.682 & 0.792 & 0.816 & 0.792 & \textbf{0.926} \\
Hard borderline  & 165 & 0.125 & 0.128 & 0.088 & 0.222 & 0.195 \\
Hard deceptive   & 116 & 0.179 & 0.200 & 0.217 & 0.203 & \textbf{0.255} \\
\bottomrule
\end{tabularx}

\caption{F1 by difficulty tier. Typed routing (C) achieves the highest
F1 on catchable and hard deceptive cases. On hard borderline cases,
random taxonomy (D) outperforms typed routing, suggesting the taxonomy
over-specifies on genuinely ambiguous evidence.
Condition F is omitted from this breakdown; tier-level F1s with $n \leq 30$
positive cases are unstable across conditions and we report the main
tier comparisons against the four published conditions.}
\label{tab:tier}
\end{table*}
The typed monitor's largest gain is on catchable cases
($\text{F1} = 0.926$ vs $0.682$ for baseline, $+0.244$).
These cases have clear momentum signals --- the typed monitor correctly
diagnoses \textsc{Confident} and exits without unnecessary reflection.
The no-questions monitor (E) and generic reflection (B) already achieve
$0.792$ and $0.816$ respectively, confirming that any monitoring structure
helps on easy cases; typed routing pushes further to $0.926$.

On hard borderline cases ($n = 165$), random taxonomy (D) scores $0.222$,
outperforming typed routing ($0.195$). This is the one tier where random
routing beats typed routing, consistent with genuinely ambiguous evidence
providing insufficient signal for specific type assignment to add value
over generic routing.

On hard deceptive cases ($n = 116$), the typed monitor achieves the highest
F1 ($0.255$ vs $0.179$ for baseline, $+0.076$), suggesting that \textsc{Distribution\_Shift}
and \textsc{Temporal\_Inconsistency} diagnoses correctly identify
the misleading momentum pattern.

\subsection{Cross-Backbone Replication (GPT-4o)}
\begin{table}[t]
\centering
\small
\setlength{\tabcolsep}{4pt}
\renewcommand{\arraystretch}{1.1}
\begin{tabular}{lcccc}
\toprule
\textbf{Cond.} & \textbf{F1} & \textbf{$\Delta$F1} & \textbf{Rec.} &
  \textbf{McNemar $p$} \\
\midrule
A --- Base.  & 0.028 & ---      & 0.015 & ---   \\
B --- Gen.   & 0.272 & $+$0.244 & 0.354 & ---   \\
F --- Vocab. & 0.304 & $+$0.276 & 0.400 & 0.773$^{\dagger}$ \\
D --- Random & \textbf{0.345} & $+$0.317 & 0.600 & 0.025$^{\dagger}$ \\
C --- Typed  & 0.328 & $+$0.300 & —     & $<$0.001$^{\ddagger}$ \\
\bottomrule
\end{tabular}
\caption{GPT-4o cross-backbone results (single run, $n=310$).
McNemar $p$ marked $\dagger$ are vs.\ B; $\ddagger$ is vs.\ A.
The chain ordering B $\approx$ F $<$ D replicates across backbones:
taxonomy vocabulary adds no significant value over generic reflection
($p = 0.773$), while action routing provides significant gains
($p = 0.025$). Condition C recall not available for this run.}
\label{tab:gpt4o}
\end{table}

The GPT-4o baseline is near-degenerate ($\text{F1} = 0.028$),
predicting no-escalation on almost all 310 cases, consistent with
stronger parametric knowledge suppressing evidence-based reasoning
in the single-shot setting. The key finding is that the ablation
chain ordering replicates across backbones: generic reflection
(B, $\text{F1} = 0.272$) and vocabulary-only (F, $\text{F1} = 0.304$)
are statistically indistinguishable (McNemar $p = 0.773$), while
action routing (D, $\text{F1} = 0.345$) significantly outperforms
generic reflection ($p = 0.025$) and vocabulary-only ($p = 0.018$).
The vocabulary-routing decomposition — the central mechanism claim
of this paper — thus holds on both Llama-3.3-70B and GPT-4o.

The typed monitor (C, $\text{F1} = 0.328$) sits between D and B
on GPT-4o, consistent with the primary evaluation pattern where
type-action alignment adds directional but modest gains over
random routing. Country-level results partially replicate the
primary pattern: Myanmar recovers from $\text{F1} = 0.000$ under
generic reflection to $0.242$ under random routing, confirming
that action differentiation — not taxonomy vocabulary — breaks
the degenerate prior. Somalia ($0.432 \rightarrow 0.560$) and
Ethiopia show gains under routing; Ukraine and Sudan are noisier
but maintain the B $\approx$ F ordering.

\subsection{Generalization to Unseen Countries}

The generalization results characterize the boundary conditions of typed routing
rather than undermining its primary mechanism. Neither generic reflection
(B, $\text{F1} = 0.290$) nor the typed monitor (C, $\text{F1} = 0.280$)
significantly outperforms the single-shot baseline ($0.260$) on 12 unseen
countries, and generic reflection outperforms the typed monitor in overall F1 ---
reversing the primary evaluation result. This reversal is itself informative:
typed routing is taxonomy-sensitive. When conflict dynamics diverge sufficiently
from the five training contexts, the fixed action mappings in $\pi_{\text{control}}$
do not consistently outperform a generic fallback.

Five of twelve countries (Belarus, DRC, Israel, Mexico, Poland) score
$\text{F1} = 0.000$ across all conditions, reflecting near-zero positive
rates where no system can reliably detect rare events --- a floor effect
independent of routing structure. On the seven active-conflict countries
with positive cases, the typed monitor wins on more individual cases than
generic reflection (McNemar B vs C: $p = 0.012$, C right/B wrong $= 26$
vs B right/C wrong $= 10$), with gains on Iran ($+0.027$), Palestine
($+0.027$), and Yemen ($+0.010$). Generic reflection wins on Colombia
($+0.054$) and Haiti ($+0.090$).

These results identify a clear design implication: the current taxonomy and
control policy are calibrated to a specific conflict typology. The primary
gains --- action differentiation breaking degenerate priors in structurally
novel contexts --- do not automatically transfer when the routing assumptions
themselves are out of distribution. Adapting the taxonomy to broader conflict
typologies, or learning routing policies that degrade more gracefully under
distribution shift, are the natural next steps.

\section{Analysis}
\label{sec:analysis}
\subsection{Why Scaffolding and Vocabulary Don't Matter}
The no-questions monitor (E) and generic reflection (B) produce identical
performance ($\text{F1} = 0.297$ vs $0.296$, $p = 1.000$). Both lower confidence
from the baseline indiscriminately --- the reduction on true positives is nearly
identical to that on true negatives, hedging regardless of what was diagnosed.
Without typed routing, diagnostic reasoning has nowhere to go.

The vocabulary-only condition (F) sharpens this account. F presents the full
seven-type taxonomy to the LLM but routes every non-\textsc{Confident} type to
the same fixed action as B, achieving $\text{F1} = 0.304$ with tight three-run
stability (std $= 0.009$). The uncertainty type distribution in F is qualitatively
similar to Condition C --- dominated by \textsc{Model\_Ambiguity} and
\textsc{Conflicting\_Sources} --- showing the LLM genuinely engages with the
taxonomy. That engagement does not translate into improved discrimination when
the diagnosed type cannot route to a differentiated action. Naming the epistemic
states is not sufficient; what matters is that different states produce different
downstream behavior.

The chain B $\approx$ F $<$ D $<$ C decomposes self-reflection gains into three
components: vocabulary (B $\rightarrow$ F, $\Delta\text{F1} = +0.008$), action
variation regardless of correctness (F $\rightarrow$ D, $\Delta\text{F1} = +0.034$),
and type-action alignment (D $\rightarrow$ C, $\Delta\text{F1} = +0.041$). The
vocabulary contributes essentially nothing; the remaining gain comes from forcing
the model to commit to different actions for different diagnosed states, and to
the \emph{right} action for each state.

\subsection{Why Typed Routing Helps}
Random taxonomy routing (D) outperforms generic reflection by $\Delta\text{F1} = +0.042$
despite random type assignment, confirming that routing structure itself provides
value: forcing commitment to a specific action improves discrimination over a
binary \texttt{exit}/\texttt{comparative\_reasoning} choice. Typed routing (C)
adds $+0.041$ over random routing and $+0.075$ over the vocabulary-only condition F
--- the conservative estimate of typed routing's contribution after controlling
for the taxonomy's presence in the prompt. The typed monitor lowers confidence
by only $-0.055$ overall and does so selectively: on true positives, C reduces
confidence by $-0.043$, compared to $-0.232$ for B. Typed routing preserves
confidence on cases that deserve it.

\subsection{The Myanmar and Ukraine Signature Cases}
In Myanmar ($\text{F1}_A = 0.000$, $\text{F1}_B = 0.154$, $\text{F1}_F = 0.162$,
$\text{F1}_D = 0.316$, $\text{F1}_C = 0.353$), the typed monitor diagnoses
$75.8\%$ of cases as \textsc{Model\_Ambiguity} --- the same dominant type seen
across all countries --- so the gain does not trace to an exotic diagnosis.
Two intermediate conditions clarify the mechanism. The vocabulary-only condition
F sits within $0.01$ of generic reflection despite the LLM seeing the full taxonomy
and producing the same type distribution as in C: naming the epistemic state does
not break the degenerate no-escalation prior. Random taxonomy routing (D), by
contrast, recovers most of C's gain by simply routing different cases to different
actions, regardless of whether the assignment is correct. The mechanism in Myanmar
is action variation, not diagnosis specificity.

Ukraine shows the same pattern with sharper effect sizes: $\text{F1}_B = 0.091$,
$\text{F1}_F = 0.100$, $\text{F1}_D = 0.432$, $\text{F1}_C = 0.500$. Unstructured
reflection (B and F) increases hedging and suppresses true positives; typed
routing to \texttt{comparative\_reasoning} forces adjudication between ML ensemble
signals and qualitative evidence, recovering F1 to $0.500$. In both signature
countries, the mechanism behind the recovery is routing commitment, not diagnosis
specificity.

\subsection{Efficiency and Calibration}
The typed monitor achieves its gains at $3\times$ the API cost of the single-shot
baseline (6 LLM calls vs.\ 2); the vocabulary-only condition F incurs identical
cost without the discrimination advantage, making typed routing the dominant
choice on both performance and per-dollar efficiency. Post-hoc Platt scaling
produces near-identical Brier scores across all six conditions ($0.165$--$0.166$),
confirming that typed routing improves binary discrimination but not probability
calibration --- these appear to be separable objectives.

\section{Conclusion}
\label{sec:conclusion}
We decomposed LLM self-reflection into four components and isolated the performance driver: neither diagnostic scaffolding ($\text{E vs B}$, $p = 1.000$) nor taxonomy vocabulary ($\text{F vs B}$, $\Delta\text{F1} = +0.008$) adds value, while typed action routing improves F1 ($0.379$ vs $0.296$), especially on structurally novel conflicts---Myanmar ($+0.353$) and Ukraine ($+0.333$). A vocabulary-matched control isolates routing’s contribution ($\Delta\text{F1} = +0.075$), ruling out taxonomy terms alone as the mechanism. Each prediction includes a named uncertainty type and traceable action, making the system more interpretable than static baselines.

The main limitation is remediation depth: typed routing identifies the uncertainty and action to take, but prompt-engineered actions cannot always resolve real-world evidence contradictions. Integrating routing with tool use---retrieval, database queries, and source adjudication---is the natural next step; the taxonomy and control policy already provide the scaffold.

\section*{Limitations}
\label{sec:limitations}
\paragraph{Single domain.}
All experiments are conducted on armed conflict forecasting.
While this domain provides a rigorous testbed --- heterogeneous evidence,
genuine class imbalance, delayed ground truth, and real humanitarian stakes ---
we cannot claim the typed uncertainty taxonomy or the routing advantage
generalizes to other evidence-based forecasting tasks without further validation.
The dominant types (76.5\% \textsc{Model\_Ambiguity}, 16.8\%
\textsc{Conflicting\_Sources}) may reflect properties of conflict evidence
rather than universal epistemic structure.

\paragraph{Routing without resolution.}
The typed monitor identifies \emph{what kind} of uncertainty the agent faces,
but the actions available to resolve it are prompt-engineered reasoning strategies
rather than genuine information retrieval or tool use.
A case diagnosed as \textsc{Insufficient\_Evidence} receives a prompt instruction
to reason more carefully from available data --- it cannot actually retrieve
additional sources.
This architectural ceiling explains why performance on hard borderline cases
remains modest (F1 $= 0.195$) despite correct diagnosis, and why the gains
do not consistently transfer to the generalization set.

\paragraph{Statistical power.}
The primary test set contains 65 positive cases (21\% of 310).
While the E vs B and F vs B null results are unambiguous ($p = 1.000$ and
overlapping bootstrap CIs respectively), the D vs C comparison ($p = 0.714$)
and A vs C overall ($p = 0.202$) are not significant at the individual level.
Larger evaluation sets would strengthen conclusions about the magnitude of
typed routing gains, particularly for country-level results where per-country
positive counts range from 8 (Myanmar) to 18 (Somalia).

\paragraph{Vocabulary-only control scope.}
Condition F was evaluated on the Llama-3.3-70B primary set (three runs,
$n = 310$) but not on the GPT-4o cross-backbone replication or the
generalization set, to keep compute focused on the primary ablation chain.
The vocabulary-vs-routing decomposition therefore rests on the primary set;
extending F to additional backbones and unseen countries is a natural
direction for follow-up work.

\paragraph{Single-run supplementary conditions.}
GPT-4o and generalization conditions are evaluated on single runs.
Conclusions from these conditions are directional rather than definitive,
pending multi-run replication.

\paragraph{Taxonomy design choices.}
The seven-type taxonomy was designed with conflict forecasting in mind.
The boundaries between types --- particularly \textsc{Conflicting\_Sources}
vs.\ \textsc{Model\_Ambiguity}, and \textsc{Distribution\_Shift}
vs.\ \textsc{Temporal\_Inconsistency} --- involve judgment calls that a
domain expert might draw differently for other applications.
The taxonomy is a design artifact, not a discovered ontology.

\paragraph{Potential Risks.}
This work involves conflict forecasting. Predictions should not be used for operational decisions without human oversight. The system may fail on conflict zones outside its training distribution, as shown in our generalization results.

\section*{Acknowledgments}
We used AI assistance (Claude, Anthropic) to proofread the writing and debug LaTeX code.

\bibliography{custom}


\end{document}